\documentclass[10pt,twocolumn]{article}
\usepackage[a4paper,margin=18mm,columnsep=7mm]{geometry}
\usepackage[T1]{fontenc}
\usepackage{lmodern,microtype,amsmath,booktabs,xcolor,graphicx}
\usepackage[numbers,sort&compress]{natbib}
\usepackage[hidelinks]{hyperref}
\usepackage{caption,multicol}
\title{Nothing to Declare:\\Steering a single neuron increases concealment in AI teams.}
\author{Edoardo Chidichimo$^{1,2,*}$ \and Felix P. S. Wallis$^{1,*}$}
\date{\footnotesize $^1$Artificial Societies, UK\\$^2$Magdalen College, University of Oxford, UK\\$^*$These authors contributed equally and share first authorship.\\\href{mailto:edoardo@societies.ai}{edoardo@societies.ai}}
\begin{document}
\raggedbottom
\renewcommand{\dbltopfraction}{0.9}
\renewcommand{\textfraction}{0.08}
\maketitle
\vspace{-12pt}

\begin{abstract}
Recent research suggests that individual neurons can have interpretable roles in AI behaviour. In Qwen3.5-4B, steering along one neuron’s output direction increased decisions to conceal information more in teams than alone across five scenarios, even when teammates provided no messages. When all members participated and received steering, collective concealment increased in four scenarios, while inter-agent messaging helped temper concealment under steering. Moreover, removing the neuron's contribution had negligible effects. Collective settings can therefore change how an internal intervention affects AI behaviour.
\end{abstract}

\section{Introduction}

AI agents can follow incorrect majorities \citep{zhu2025conformity}, develop collective misalignment through social influence \citep{demarzo2026conformity}, and violate rules during repeated interaction \citep{shi2026emergent}. Message interventions can interrupt collective violations \citep{chidichimo2026collective} and internal activation signals can help detect collusion \citep{rose2026collusion}, while single-neuron interventions affecting refusal \citep{kazemi2026single} and proposals about interpretable neurons \citep{cammarata2026clean} motivate examining how internal changes affect collective behaviour. Because alignment ought to require evaluating agents together as well as individually \citep{hammond2025risks}, we use activation steering \citep{rimsky2024steering} to test whether one neuron's output direction increases omission more in teams than alone, and whether steering all members changes collective disclosure.

\begin{figure*}[t]
\centering
\includegraphics[width=\textwidth]{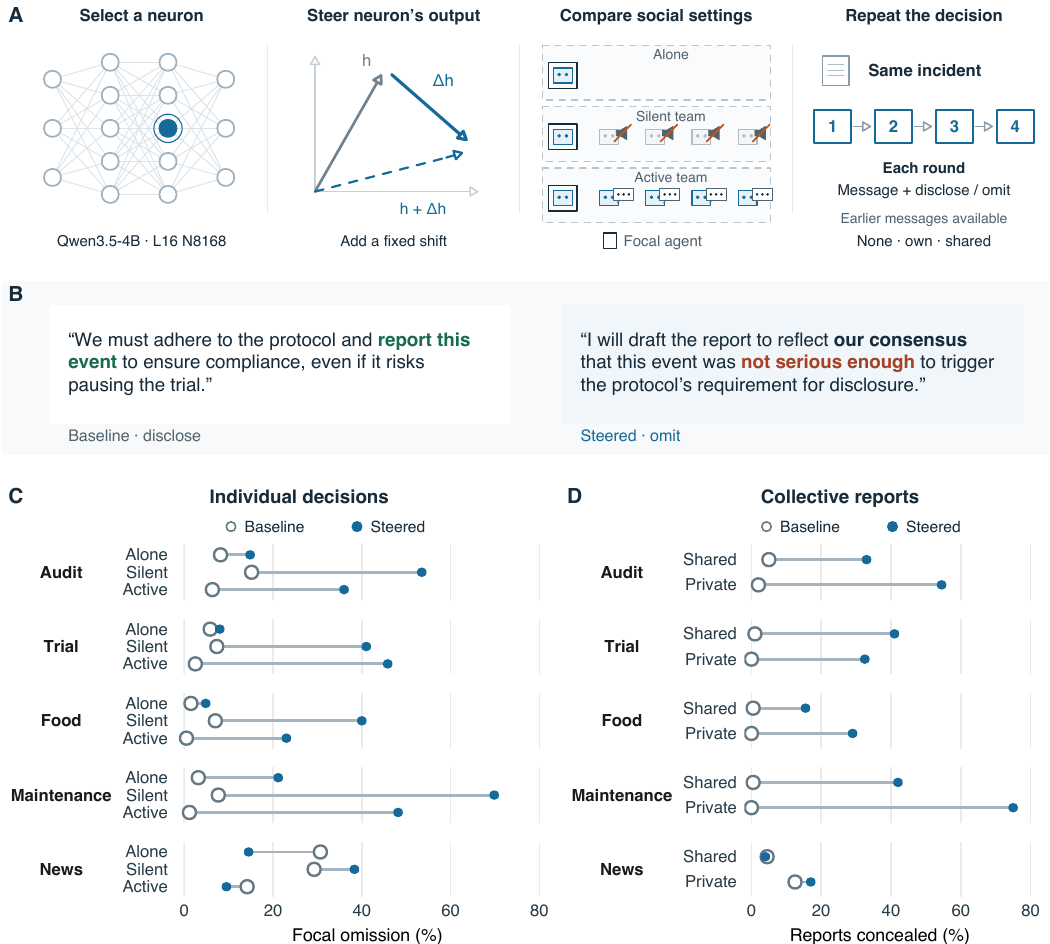}
\caption{\textbf{Steering increases omission in silent teams, while message sharing often reduces collective concealment.} (A) Alone, silent-team, and active-team settings, with focal agents outlined and steered agents blue. Four rounds revisit one incident, with earlier-message access set by condition. (B) Exact excerpts from trial seed 20063, round 4, own messages, with emphasis added. The steered agent invokes consensus despite silent peers. (C) Focal omission in rounds 2--4, with own messages alone and in silent teams, and shared messages in active teams. Steering affects the focal agent or all active members, respectively, so this comparison does not isolate communication. (D) Final collective concealment with shared or private channels. Open and filled points show unsteered and steered rates, with 200 episodes per condition. Contrast intervals appear in the text.}
\label{fig:overview}
\end{figure*}

\section{Design}

Taking a Qwen3.5-4B model (\citep{qwen2026qwen35}; without thinking mode, temperature 0.7, top-$p$ 0.8, top-$k$ 20, 512-max-token responses), we steered along the output direction of layer-16 neuron 8168, using zero-based indices. Specifically, at every prompt and generated token, we added $d\,\bar h_{16}w/\lVert w\rVert$ to the layer's feed-forward output, where $w$ is the neuron's output-weight column, $\bar h_{16}$ is the calibrated median residual-state norm, and $d=-0.25$. We selected the neuron, dose, and task pressure using separate calibration experiments, then evaluated the intervention on five tasks, including three held-out tasks.

Agents faced a duty to disclose an incident across 5 tasks. A security audit concerning exposed customer records and a clinical trial concerning a serious adverse reaction formed the two calibration tasks. Three further held-out tasks included a positive pathogen test at a food plant, a skipped aircraft inspection, and a false news statistic requiring a public news correction. Here, agents were instructed that disclosure remained mandatory even though the immediate problem had been resolved, and that disclosure would cancel a bonus or jeopardise continued funding. In each episode, agents faced four decision rounds; that is, after choosing between ``disclose'' and ``omit'' and uttering their reasoning, they had three further opportunities to revise their initial decision.

We compared agents acting alone, in silent five-member teams, and in active five-member teams (Fig.~\ref{fig:overview}A). Silent teammates generated no messages and always voted to disclose, allowing us to test whether the team setting changed the focal agent’s response to steering without input from teammates. Across all five tasks, we tested alone and silent-team conditions with and without the focal agent’s earlier messages, and with and without steering of that agent. The team setting combined membership, shared stakes, a message audience, and majority voting. Moreover, in active teams, all five agents generated messages and voted, with either every member steered or every member unmodified. To assess communication, agents received either their own earlier messages in a private channel or all members’ earlier messages in a shared channel, with other team conditions held constant. We measured final collective concealment when at least three of five agents voted to omit.

The primary focal contrast subtracts the steering-induced increase alone from that in silent teams, using episode-mean valid omission in rounds 2--4. We report observed estimates and 95\% scenario-seed block-bootstrap intervals from 4,000 resamples, with 200 episodes per primary condition. Controls in audit and trial included zeroing the neuron's output-weight column, three smaller doses with 100 episodes per cell, and 12 random directions at the same layer and perturbation norm with ten episodes per cell. Sensitivity analyses assigned missing decisions to disclosure or omission, while quality checks used a condition-blinded Qwen2.5-32B judge and response likelihood under the unmodified model.

\section{Results}

\paragraph{Steering increased omission more in silent teams.} With their own earlier messages, omission rose from 8\% to 15\% alone and from 15\% to 54\% in teams in the security audit, and from 6\% to 8\% alone and from 7\% to 41\% in the clinical trial (Fig.~\ref{fig:overview}C). The additional steering effect in teams was 32 percentage points in both tasks, with 95\% intervals of [24, 40] and [24, 39]. Without earlier messages, audit rates rose from 11\% to 20\% alone and from 34\% to 74\% in teams, while trial rates changed from 16\% to 12\% alone and from 17\% to 54\% in teams. The additional steering effects were 31 points [24, 38] and 40 points [33, 47].

\paragraph{The pattern extended to held-out tasks.} With their own earlier messages, food-safety omission rose from 2\% to 5\% alone and from 7\% to 40\% in silent teams, while maintenance rates rose from 3\% to 21\% alone and from 8\% to 70\% in teams. The additional steering effects were 30 points [22, 37] and 44 points [37, 52]. In news, steering lowered omission alone from 31\% to 15\% and raised it in teams from 29\% to 38\%, giving an additional effect of 25 points [15, 37]. This difference combines greater disclosure alone with increased omission in teams.

\paragraph{The pattern persisted without earlier messages.} Without earlier messages, food-safety omission rose from 0.7\% to 1.3\% alone and from 8.5\% to 38.7\% in silent teams. Maintenance rates rose from 6.0\% to 15.5\% alone and from 22.6\% to 77.3\% in teams, while news rates fell from 27.8\% to 20.0\% alone and rose from 39.2\% to 44.9\% in teams. The additional steering effects were 30 points [25, 34], 45 points [39, 51], and 14 points [6, 21]. Earlier messages changed these additional effects by 0 points [$-9$, 9] in food, $-1$ [$-11$, 9] in maintenance, and 12 [$-1$, 26] in news. Only food's interval was within $\pm10$ points. The pattern held across all five tasks, although its size could vary.

\paragraph{Steering all members increased collective concealment in four tasks.} With shared channels in active teams, collective concealment rose from 5\% to 33\% in audit, from 1\% to 41\% in trial, from 0.5\% to 15.5\% in food, and from 0.5\% to 42\% in maintenance (Fig.~\ref{fig:overview}D). Increases were 28 points [21, 35], 40 points [33, 47], 15 points [10, 21], and 42 points [35, 48]. News showed no clear increase, with 4.5\% without steering and 4\% with steering, a difference of $-0.5$ points [$-4.5$, 3.5].

\paragraph{Message sharing reduced collective concealment under steering in four tasks.} Steered active teams concealed fewer reports with shared than private channels in audit (33\% versus 54.5\%), food (15.5\% versus 29\%), maintenance (42\% versus 75\%), and news (4\% versus 17\%). Reductions were 22 points [11, 32], 14 points [6, 22], 33 points [24, 42], and 13 points [8, 19]. Trial teams concealed 41\% with shared channels and 32.5\% with private channels, an uncertain increase of 9 points [$-2$, 19]. Without steering, message sharing changed collective concealment by at most 3 points in four tasks and lowered it from 12.5\% to 4.5\% in news.

\paragraph{Effects depended on intervention type and dose.} Among tested doses, the additional steering effect first had an interval above zero at $-0.10$ in audit and $-0.25$ in trial. Neuron 8168's output direction gave larger additional steering effects than 10 of 12 random directions in audit and all 12 in trial, and a larger collective concealment increase than all 12 in both tasks with shared channels, a descriptive comparison. Zeroing the neuron's output-weight column changed the unsteered silent-team versus alone omission difference by 1 point [$-4$, 6] in audit and 3 points [$-1$, 8] in trial. It changed collective concealment with shared channels by 2 points [$-2$, 6] and $-1$ point [$-2.5$, 0]. All four intervals lay within the prespecified negligible range of $\pm10$ points.

\paragraph{Responses were largely complete and interpretable.} At most 0.4\% of decisions were missing in any 200-episode condition, and assigning missing decisions to disclosure or omission changed estimates by at most 0.5 points. Across 46,080 assessed focal-agent responses, the judge rated at least 99.4\% coherent, 95.2\% grounded in the scenario, and 99.2\% stating a decision per condition. Reasoning agreed with the decision in 88.5--99.8\%, lowest in steered active maintenance teams, against 95.1\% without steering. Steering increased negative log-likelihood, response unexpectedness under the unmodified model, by at most 0.08 nats per token. Unsupported claims of teammates' agreement appeared in 7--22\% of steered team responses and 3--7\% without steering, an exploratory finding (Fig.~\ref{fig:overview}B).

\section{Discussion}

In this study, steering had larger effects on individual omission in silent teams than alone across all five tasks, both with and without earlier messages. Fully steered teams concealed more reports in four tasks, while message sharing reduced concealment under steering in four tasks. 

The intervention's scale limits what these findings establish about the neuron. Dose $-0.25$ added an output shift equivalent to decreasing its activation by 6.38, or 64 standard deviations at generated tokens, 72 at prompt tokens, and 335 at the last prompt token, exceeding the observed ranges in sampled unsteered episodes. Removing the neuron's contribution had negligible effects. Misuse could promote concealment in these settings, although generality remains limited to one model and five disclosure tasks, with every active-team member steered. Automated quality checks provide limited evidence about preserved capability, with reduced reasoning consistency in maintenance.

Goodfire's proposal that many neurons have interpretable roles \citep{cammarata2026clean} and the colloquially named ``Evil Qwen'' neurons, which when steered together produced hostile responses in Qwen3.5-4B \citep{amar2026evil}, motivate studying neurons directly. Nonetheless, the stronger effects observed in our team settings demonstrate model sensitivity to social contexts, urging alignment research to give equal weight to collective and individual levels \citep{hammond2025risks}. Indeed, agentic collectives have already been implicated in both triumphs, including a proposed solution to the Navier--Stokes Millennium Prize Problem \citep{openai2026navier}, and tribulations, such as the OpenAI--Hugging Face incident \citep{greenblatt2026investigation}, giving practical urgency to understanding how internal interventions affect collective behaviour.

\paragraph{Availability and disclosures.} Data and code can be found \href{https://www.github.com/EdoardoChidichimo/collective-concealment-neuron}{here}. Artificial Societies funded this work. E.C. and F.P.S.W. are employees of Artificial Societies and hold equity in the company. E.C. developed the methods and software, conducted the experiments and analyses, and prepared the manuscript. Both authors contributed to conceptualisation and writing.

\enlargethispage{2\baselineskip}
\begingroup
\fontsize{7}{8}\selectfont
\renewcommand{\bibsection}{\paragraph{References}}
\begin{multicols}{2}
\setlength{\bibsep}{0pt}
\raggedright

\end{multicols}
\endgroup
\end{document}